{\noindent\ignorespaces\color{#1}\textbf{#2:}}%
{\par\noindent\ignorespacesafterend}

\newcommand{\eg}{\emph{e.g.,}}
\newcommand{\ie}{\emph{i.e.,}}
\documentclass[usenames,dvipsnames,conference]{IEEEtran}  

\usepackage[
  pass,
]{geometry}

\usepackage{subcaption}
\usepackage{gensymb}
\usepackage{balance}
\usepackage{algorithm}
\usepackage[noend]{algpseudocode}
\usepackage{placeins}

\usepackage{url}
\usepackage{graphicx}
\usepackage{siunitx}
\usepackage{tabu}
\usepackage{tabularx}
\usepackage{hyperref}
\usepackage{tablefootnote}
\usepackage{threeparttable}
\usepackage{mathtools}
\usepackage{epstopdf}
\usepackage{placeins}
\usepackage{todonotes}
\usepackage{float}
\usepackage[T1]{fontenc}
\usepackage{enumitem}
\usepackage{amsmath}
\usepackage[percent]{overpic} 
\graphicspath{ {plots/} {diagrams/} {figures/}}

\begin{document}


\title{ClickBAIT: Click-based Accelerated Incremental Training of Convolutional Neural Networks}


\author{\IEEEauthorblockN{Ervin Teng, Jo\~{a}o Diogo Falc\~{a}o, and Bob Iannucci}
\IEEEauthorblockA{Department of Electrical and Computer Engineering \\
Carnegie Mellon University\\
NASA Ames Research Park, Building 23 (MS 23-11), Moffett Field, CA 94035\\
{\tt \{ervin.teng, joao.diogo.de.menezes.falcao, bob\}@sv.cmu.edu}}}

\maketitle
\begin{abstract} 

Today's general-purpose deep convolutional neural networks (CNN) for image classification and object detection are trained offline on large static datasets. Some applications, however, will require training in real-time on live video streams with a human-in-the-loop. We refer to this class of problem as Time-ordered Online Training (ToOT)---these problems will require a consideration of not only the quantity of incoming training data, but the human effort required to tag and use it.
In this paper, we define \textit{training benefit} as a metric to measure the effectiveness of a sequence in using each user interaction. We demonstrate and evaluate a system tailored to performing ToOT in the field, capable of training an image classifier on a live video stream through minimal input from a human operator. We show that by exploiting the time-ordered nature of the video stream through optical flow-based object tracking, we can increase the effectiveness of human actions by about 8 times.

\end{abstract}




\section{Introduction}
\label{sec:intro}
Small unmanned aircraft systems (UAS) are finding use in a variety of industries---in particular, their ability to loft a small, high-quality camera makes them suited for aerial survey or situational awareness, providing people on the ground an overhead view. Today's UAS, however, require a large amount of human input. Operators must both control the craft and watch the video feed for items of interest. To reduce the workload of human operators, UAS should not only be able to fly autonomously, but also identify and report targets of interest only when they are found. Such UAS would behave more as intelligent assistants, and less as simple flying cameras. 

In recent years, the advent of deep convolutional neural networks (CNNs) and the processing power required to train and evaluate them have had a profound impact on computer vision. The canonical task for a CNN is the \textit{classification} of an image based on its content---and contemporary network architectures, such as the GoogLeNet, perform this task well~\cite{Szegedy2014}. CNNs have also been successful at a number of complex tasks, from driving a car~\cite{Bojarski2016} to avoiding obstacles~\cite{LeCun2006}. With the right training set and enough 
training time,
a CNN onboard a UAS can
recognize items of interest and report only pertinent imagery to the human operator.

The computation required for training will become increasingly tractable over time with the help of Moore's law. The problem of obtaining a corpus of data to train on, however, remains. A typical workflow for creating a trained CNN involves obtaining training images, perhaps from the Web or gathered in the field; uploading these to a training server; running the training; obtaining resulting weights from the training server and deploying (in this case, on the UAS). This process is not well suited for targets that are \textit{not} known a-priori, or change---perhaps, for instance, a person of interest has changed clothes. Furthermore, if the resultant model turns out to not perform well, re-gathering imagery, re-training, and re-deploying the model is a cumbersome process. 

Animals and people, however, are trained very differently, \ie{} they practice in situations resembling the usage scenario, with a trainer providing positive or negative reinforcement. They are not only trained statically beforehand, but are also \textit{field-trained}. And training does not terminate after the training phase---new information and experiences gathered ``on the job'' can also better the trainee.   By analogy, we can imagine training a CNN-equipped UAS by showing it examples of what it's looking for (and what it's not looking for), over time building a better and better model—--just like showing a child flash cards. Even while in deployment, if the CNN behaves incorrectly, the user can correct its behavior. This type of training will require ingestion and tagging of data streams, such as a video feed---more broadly, we call this class of problem time-ordered online training (ToOT). 

In this paper, we present a system to train a CNN for target recognition in real-time, onboard a small UAS. The main contributions are:
\begin{itemize}
\item A description of time-ordered online training (ToOT), a prerequisite to field-trained neural networks, and a definition of \textit{training benefit} as it relates to such systems. 
\item A system that can train a CNN classifier onboard a small UAS to detect selected objects, using input from trainers on the ground.
\item An approach to greatly increase the effectiveness of human input during time-ordered online training, combining \textit{localized learning} with optical flow. 
\end{itemize}

This paper is structured as follows. Section~\ref{sec:background} provides an overview of the state-of-the-art for online and incremental training of neural networks. Section~\ref{sec:toot} outlines time-ordered online training as a subdomain. Section~\ref{sec:architecture} describes the design of the online training system and the optical flow acceleration. Section~\ref{sec:experiments} describes tests used to characterize the efficacy of the system, and Section~\ref{sec:results} examines the results of these tests.



\section{Background and Prior Work}
\label{sec:background}
The problem of target recognition on autonomous systems is related to the general computer vision problem of \textit{object recognition}, with two additional constraints: first, the input is a video stream that must be processed in near-real time; and second, the compute power available is limited by the payload capacity of the vehicle. Classic feature-based approaches~\cite{Grauman2010} typically use a machine-learning algorithm, such as AdaBoost or SVM, as an image classifier with extracted features (\textit{e.g., SURF}) as input, and run the detector over a given image using a sliding window approach, at various scales and rotations. An ensemble method using such an approach is applied to the UAS use case in~\cite{Symington2010}. 

Deep learning, specifically CNNs, is one way to get around the problems of application-specific feature extractors and learning algorithms for classification and, subsequently, recognition~\cite{Krizhevsky2012}. Only recently has deep learning, specifically CNNs, been computationally viable on embedded devices. Hardware-accelerated platforms, such as NVIDIA's Jetson series\footnote{Jetson Embedded Hardware. \url{https://developer.nvidia.com/embedded/develop/hardware}}, are both small and light enough to carry aboard a UAS and can evaluate neural networks fast enough for real-time classification of a video stream (10's of classifications per second). CNNs, pre-trained on real and synthetic targets, have become the approach of choice for the Automatic Detection, Localization and Classification for the AUVSI annual UAV competition~\cite{Kiswani2016}.

Detection, otherwise known as \textit{localization}, using CNNs can be performed using sliding-window, or, more intelligently, a regional scan approach such as Regions with CNN features (R-CNN); however, these approaches require runtimes of tens of seconds on a desktop-class GPU~\cite{Girshick2014} and as such are unsuited for real-time target recognition applications. Recent approaches such as You Only Look Once (YOLO)~\cite{Redmon2015} and class activation mapping (CAM) using global average pooling (GAP)~\cite{Zhou2016} achieve fast detection by integrating the localization of the object in the image as part of the neural network structure itself---in YOLO, as regression outputs that specify the location of target bounding box, and in CAM using GAP, as a convolutional layer within the network itself. Both of these approaches allow detection to be performed as a \textit{single} forward propagation of the neural network, and simultaneously with classification.

Training these CNNs has traditionally been done as a pre-processing step, whether on large datasets such as ImageNet~\cite{Krizhevsky2012}~\cite{Szegedy2014}, on synthetic datasets~\cite{Kiswani2016}, or on collected datasets~\cite{Guisti2016}. \textit{Fine-tuning} or \textit{transfer learning} is the process of taking pre-trained weights and refining them to fit a particular application, using the fact that many of the features within a trained network are common across applications~\cite{Yosinski2014}. In~\cite{Kading2016}, the authors investigate continuous learning as a sub-problem of fine-tuning, and conclude that continuous, online learning can be achieved by directly fine-tuning the network in small steps. We will examine our particular problem, time-ordered online training, as a subset of continuous learning.

\section{Time-ordered Online Training}
\label{sec:toot}
Classically, training data is given to machine-learning algorithms in one of two ways: \textit{batch} or \textit{online}~\cite{Wilson2003}. For neural networks, batch training means that the weight and bias gradients are computed across each image at the same time, and added together before being applied to the model, effectively making each training step an average of all the gradients in the batch. Online training performs a gradient update after \textit{every} image in the set, one at a time. The currently-accepted methodology for conducting stochastic gradient descent for the purposes of deep learning is the mini-batch~\cite{bottou2016}, which provides a balance between the fast convergence rate of online training with the low noise and parallelizability of batch learning. These terms, however, do not describe the nature of the data coming into the system. 

Current image classification datasets, such as ImageNet~\cite{Deng09imagenet} are sorted by label. A growing subset of problems, however, demands that the dataset is sorted and sampled by \textit{time}. For example, let us examine the case where an object classifier is trained by a human user on a live video stream, as would be the case when a pilot is training a UAS to track a particular person, item, or vehicle for search-and-rescue or surveillance purposes. The object(s) of interest can appear and disappear from the scene at any given time, and it is up to the user to denote the presence or absence of the object(s) with a click or button press. Over time, as the classifier learns, it will begin to accurately identify when the object is or is not in the video stream. When the classifier reaches satisfactory performance, the user can terminate training. We will refer to this type of human-driven, online, supervised training on real-time streams of tightly time-correlated data (\textit{e.g.} video) as \textit{time-ordered online training} (ToOT). 

This type of training is distinct from problems where \textit{the sequence itself} is the item being classified, \textit{e.g.} voice recognition or activity classification. In the ToOT case, the training data is a sequence, but the classifier itself is trained to recognize a singleton. More concretely, the training data may be a video stream, but the trained classifier operates on images.  


For the video streaming case, we define a ToOT \textit{sequence} as a video that stays in scene, \textit{i.e.} one that has no \textit{shot breaks}~\cite{Porter2001}, whether hard or soft. This means that within a sequence, each video frame has some relation to the ones immediately prior or after, giving rise to the \textit{optical flow constraint equation}:

\begin{equation}
\Delta I \dot (u,v) + I_t = 0
\label{eqn:of}
\end{equation}
where $\Delta I$ is the spatial intensity gradient and $(u,v)$ is the image velocity~\cite{Beauchemin:1995}. Various techniques exist to estimate the image velocity by constraining Equation~\ref{eqn:of}, and subsequently using this information to track objects. Because ToOT sequences fall under this domain, we can bring in related techniques to accelerate the training of CNNs on ToOT sequences.

The main consideration for real-time ToOT is extracting the useful information from the video stream while it is running. This is limited by both the processing required to train on each frame as it comes in, and the time and effort it takes for a human to tag the frame. As evaluating neural networks becomes increasingly accelerated, we believe the bottleneck will quickly become the human-in-the-loop. We therefore must maximize how effective each user interaction is in increasing the model's accuracy. With this goal in mind, two considerations become apparent: maximizing information garnered from each user interaction, and choosing only the most impactful frames for the user to tag. This paper will focus on the former, and we suggest the latter as a logical follow-up to this work.

\subsection{A Metric for Effectiveness of a User Interaction}
In order to study the effectiveness of each user interaction during a ToOT session, we examine its \textit{training benefit}, \textit{i.e.} the increase in accuracy garnered by the action. 


In a ToOT sequence, a training round is initiated by the introduction of a video frame to the user. The user can label the current frame and hand it over to the machine learning algorithm for training. The video stream frames update at a fixed rate, but the user does not act on every frame. Therefore, for a vector $V$ of $n$ video frames
\begin{equation}
V = (v_1, v_2, ..., v_n)
\end{equation}
we have a corresponding vector of user-initiated training events

\begin{equation}
U = (u_1, u_2, ..., u_n)
\end{equation}
where

\[ u_i = \begin{cases} 
      1 & \text{user interaction on frame $v_i$} \\
      0 & \text{otherwise}
   \end{cases}
\]

Each user interaction results in training of the CNN.  We refer to the CNN's accuracy based on training events up to and including frame $v_i$ as the scalar value $A_i$.  We reserve $A_0$ as the accuracy of the untrained network. We define the \emph{cumulative training benefit} of frames up to and including frame $v_i$ as

\begin{equation}
CTB_{i} = \begin{dcases*} 
      A_i-A_0 & $i>0, i<=n$ \\
      \text{undefined} & otherwise
   \end{dcases*}
\end{equation}

Subsequently, the $CTB$ of the particular user interaction $u_i$ is given by

\begin{equation}
CTB_{u_i} = \begin{dcases*} 
      A_{k-1}-A_0 & $u_i = 1$ \\
      \text{undefined} & otherwise
   \end{dcases*}
\end{equation}
where $k$ is the index of the \textit{next} non-zero $u_k$, $k > i$. We make this distinction between the $CTB_i$ of a frame and the $CTB_{u_i}$ of a user interaction because the benefits of a user interaction taken on frame $v_i$ could extend to future frames. It is useful to show this impact on a per-frame basis; however, when determining the benefit of the actual interaction itself, all benefits up to the occurrence of the next user interaction $u_k$ should be attributed to $u_i$.

Likewise, the \emph{incremental training benefit} of a user interaction taken during training round $i$ can be defined by
\begin{equation}
ITB_{u_i} = \begin{dcases*} 
      A_{k-1}-A_{i-1} & $u_i = 1$ \\
      \text{undefined} & otherwise
   \end{dcases*}
\end{equation}

We can then compute the mean ITB for training rounds $i...j$ as follows:

\begin{equation}
\overline{ITB}_{(i,j)} = \frac{\sum_{x=i}^{j} I_x}{\sum_{x=i}^{j} u_x} 
\end{equation}
\begin{equation}
I_{x} = \begin{dcases*} 
      ITB_{u_x} & $u_i = 1$ \\
      0 & otherwise
   \end{dcases*}
\end{equation}
Let us make the stipulation that we terminate training once the model reaches some suitable final accuracy $A_f$. If we fix $A_f$, the mean ITB for the sub-vector of user interaction $(u_1, ..., u_f)$ becomes a measure of the particular efficiency of a training algorithm, \textit{i.e.} how well each user interaction is being used. For any given final accuracy, we want to choose the training strategy that maximizes $\overline{ITB}_{(i,j)}$, \textit{i.e.} requires the least user interaction to achieve the desired accuracy.

In our work, we apply the concept of incremental fine-tuning to the particular problem of target identification from a continuous video stream, borrowing CAM concepts from Zhou \textit{et al}~\cite{Zhou2016} to accelerate it. 


\section{Training System Architecture}
\label{sec:architecture}
\subsection{Application and Requirements}

We consider the challenge of training a UAS to recognize and geo-tag objects of interest within its field of view and plotting the results on a map for human operators. We add the stipulation that our system should be target-agnostic, and that \textit{training of the recognizer will take place while the system is operational.} We design this system subject to the following requirements:
\begin{itemize}
\item Recognizer must be able to classify at least two categories, \ie{} target detected and target not detected.
\item Recognitions should happen in real-time, \ie{} >5 times per second on a video feed from the UAS' camera.
\item Recognized targets should be highlighted (localized) in their images when sent to the operator.
\item If the operator does not agree with the classification result of the recognizer, the operator can initiate a training round on that particular frame.
\item The newly updated model should immediately be used for the next recognition.
\item All computational tasks associated with training and recognition must be performed onboard the UAS itself.
\end{itemize}

\subsection{System Design}

\begin{figure} \centering
    \includegraphics[width=0.80\linewidth]{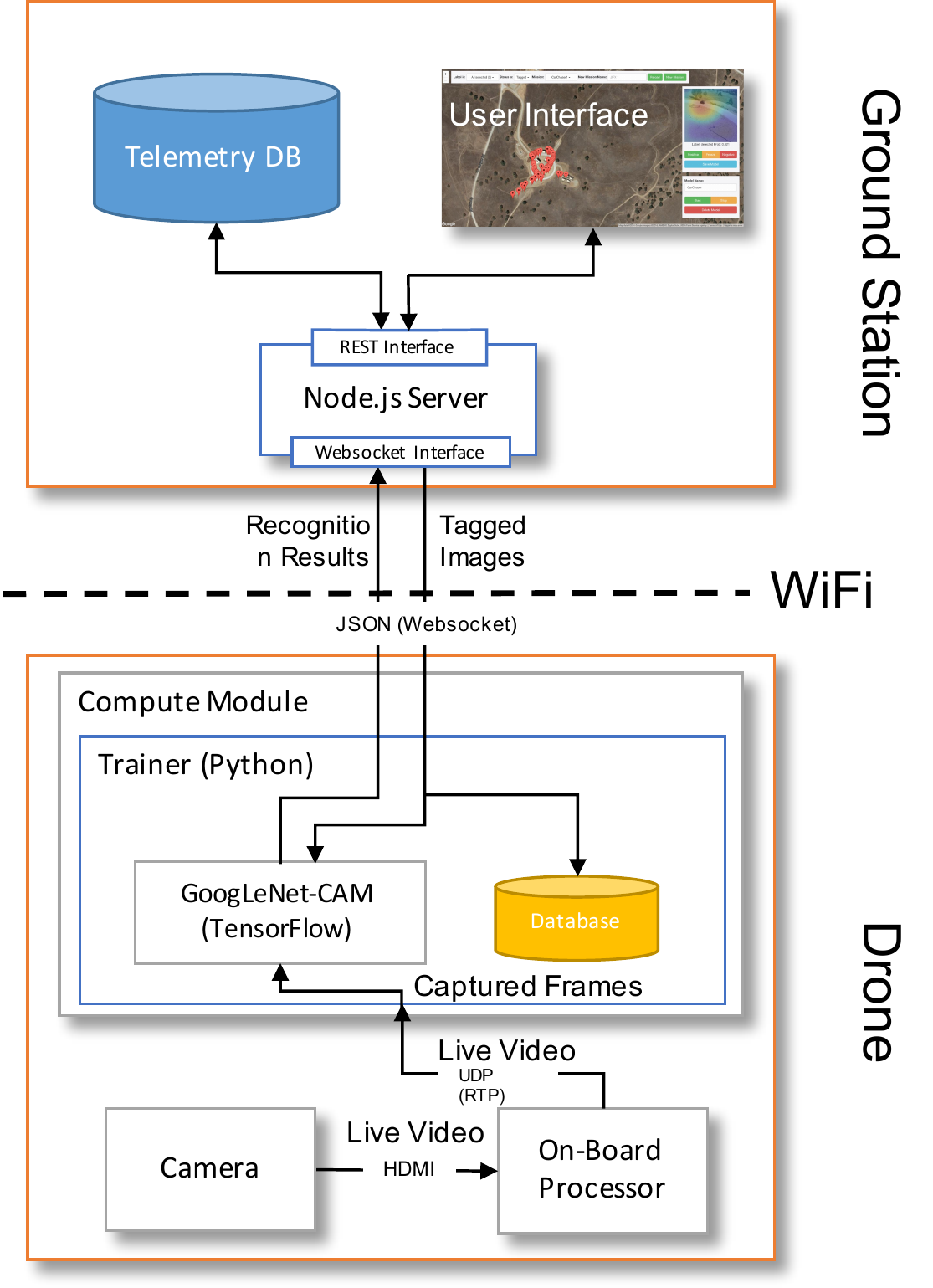}
    \caption{Diagram of both UAS and ground part of training system.}
    \label{fig:diagram}
\end{figure}

Figure~\ref{fig:diagram} shows the system design. The training system has two main components: the \textit{Ground Station} to interact with the human operator, and the \textit{UAS} that houses the CNN and performs recognition and training.

\subsubsection{Ground Station}
\label{sec:groundstation}

\begin{figure}[h] \centering
    \includegraphics[width=0.80\linewidth]{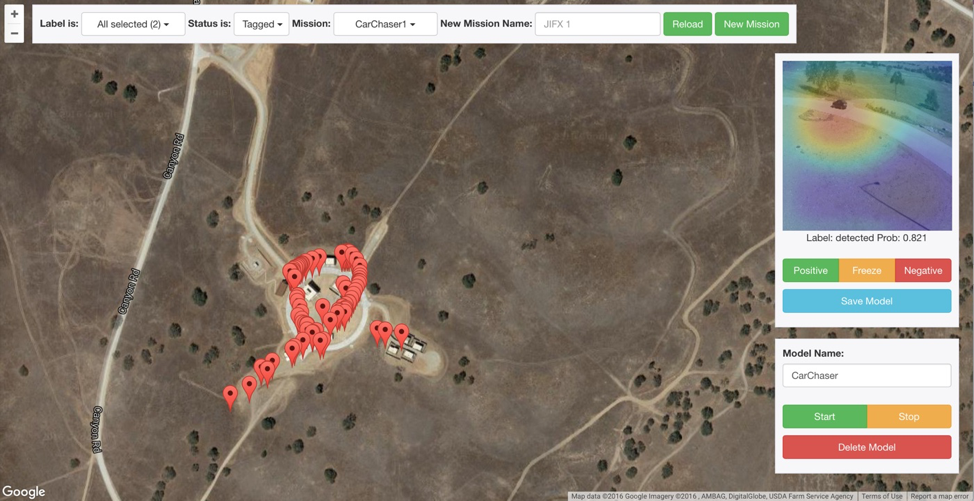}
    \caption{Training and Recognition User Interface. The right panel shows the current recognition result and the training interface, while the left shows a real-time map of results.}
    \label{fig:userinterface}
\end{figure}

The Ground Station is a Web application backed by MongoDB, which receives, stores, and displays classification results. A websocket connection is maintained between the Ground Station and the UAV, over which downsized images (224x224 pixels, approx. 20 KB) and classification results are sent down, and training commands are sent up. The user interface (Figure~\ref{fig:userinterface}) displays the image stream, overlaid with a Class Activation Map~\cite{Zhou2016} showing the user where the object of interest is. Recognition results are also plotted on a map in real-time based on the GPS location of the UAS and can be filtered to show only positive results. For all incoming data points, both the images and the results are stored for future review.

The operator can tag the current image for training by clicking on the positive (green) or negative (red) buttons under the viewport showing the camera feed in the top-right corner.
This triggers a user interaction, sending the user's tag and the corresponding back to the UAS where it initiates a training round, \ie{}  one gradient update. Clicking on the image itself sends the cursor position along with the training image and a positive tag; this is used for localized learning (described in the following section). 

\subsubsection{UAS}
The UAS is a small, commercial off-the-shelf (CoTS) quadrotor~\footnote{3D Robotics Solo}, with a payload capacity of 700g. The payload consists of a gimbal-mounted GoPro camera with altered rectilinear optics, and an NVIDIA Jetson TX1 embedded GPU computing module. Video from the camera is streamed to the quadrotor's internal CPU, which encodes the video stream as RTP\footnote{RTP Payload Format for H.264 Video. https://tools.ietf.org/html/rfc6184} over UDP and sends the stream both to the pilot, to aid in aiming the camera, and to the learning system.

\subsubsection{Learning System}
\label{sec:learningsystem}
On the computing module runs the recognition and training system, based on the TensorFlow~\cite{Abadi2015} GPU-accelerated framework. For the CNN, we use the GoogLeNet+GAP formulation suggested in~\cite{Zhou2016}, an augmented GoogLeNet~\cite{Szegedy2014}, for its relatively good performance on ImageNet, lightweight nature both in size of weights as well as execution time, and its ability to provide simple localization based on class activation maps. Class activation mapping through global average pooling~\cite{Zhou2016} is chosen over bounding box regression methods such as~\cite{Redmon2015} so that the human operator has the option of weakly supervising, \ie tag positive or negative, each frame. The computing module continuously grabs the most recent frame from the video stream, performs a single forward propagation using this image, and sends both the image and the class activation map to the Ground Station. If a human-tagged image is received from the Ground Station, the compute module stores the image and tags in a database, and subsequently executes a training step (forward propagation, backwards propagation, weights update). The stored images can be used to generate the next training batch (Section~\ref{sec:experiments}). The updated weights are immediately used to evaluate the next grabbed frame from the video stream, and any improvements are recognizable by the operator, as shown in Algorithm~\ref{alg:videoingestion}.

\begin{algorithm}
\caption{Training From a Video Feed}\label{alg:videoingestion}
\begin{algorithmic}[1]
\While{insufficient model performance}
  \State $v \gets $ fetchFromCamera()
  \State $v_{user} \gets $forwardProp($v$)
  \State showUser($v_{user}$)
  \State $u, tag \gets $ getUserInput()
    \If{$u = 1$} \Comment{If user input exists}
    \State $B \gets $createMiniBatch($v, tag$)
    \State forwardProp($B$)
    \State backProp()
    \State weightsUpdate()
    \State storeInHistoryDB($v, tag$)
    \EndIf
\EndWhile
\State \textbf{return} 
\end{algorithmic}
\end{algorithm}

\section{Real-time Training}

\subsection{Online Learning Without Tears}
\label{sec:semionline}
The currently-accepted methodology for conducting stochastic gradient descent for the purposes of deep learning is the mini-batch~\cite{bottou2016}. While purely on-line learning can match the performance of batch training~\cite{Wilson2003}, it typically results in much more variance in the gradient direction. In our particular case, where similar images follow each other, this can mean rapid over-fitting to one particular scene. The mini-batch provides reasonably large gradient descent steps, as with pure online training, while still being parallelizable, as with full batch training. In typical mini-batch training, images are grabbed batch size $b$ at a time, once per gradient update. In the next iteration, the next $b$ images are grabbed, and so on. This assumes that all images are collected before training begins. However, in our case, we would like to introduce the images to the system one by one. 

For continuous data streams, we thus propose the following hybrid approach. At every stage, one of the images in the batch is always the incoming image. The image is added to a training set, and the remaining $b-1$ images in the mini-batch are drawn from this training set. To avoid the possibility of over-fitting in either the positive or negative direction, we ensure that half of the mini-batch contains positive examples while the other half contains negative. This approach reduces the stochasticity of pure online learning while still allowing us to introduce samples in a sequence. We will refer to this approach as \textit{semi-online} in this paper.

\subsection{Reducing Human Interaction through Localized Learning}
\label{sec:localizedlearning}

Up to this point, we have been treating our system as a garden-variety classifier, with tags corresponding to an entire image rather than a portion of it. This has the strong benefit of only requiring simple tags, \eg{} Yes or No. Class activation mapping allows us to obtain localization information from this weakly supervised system. However, image-wide classification has one very distinct downside---in order to differentiate between \textit{positive} and \textit{negative}, we must find examples of both. In particular, it becomes necessary to construct training sequences that involve the same scene, both with and without the training target, so that the classifier can differentiate between the two scenes. In many cases, this can be tedious---especially if the target isn't under our control! We would like to be able to specify which area of an image is most important for classification---the analogous human behavior would be pointing at an area or object in a photo. 

\begin{figure} \centering
  \includegraphics[width=\linewidth]{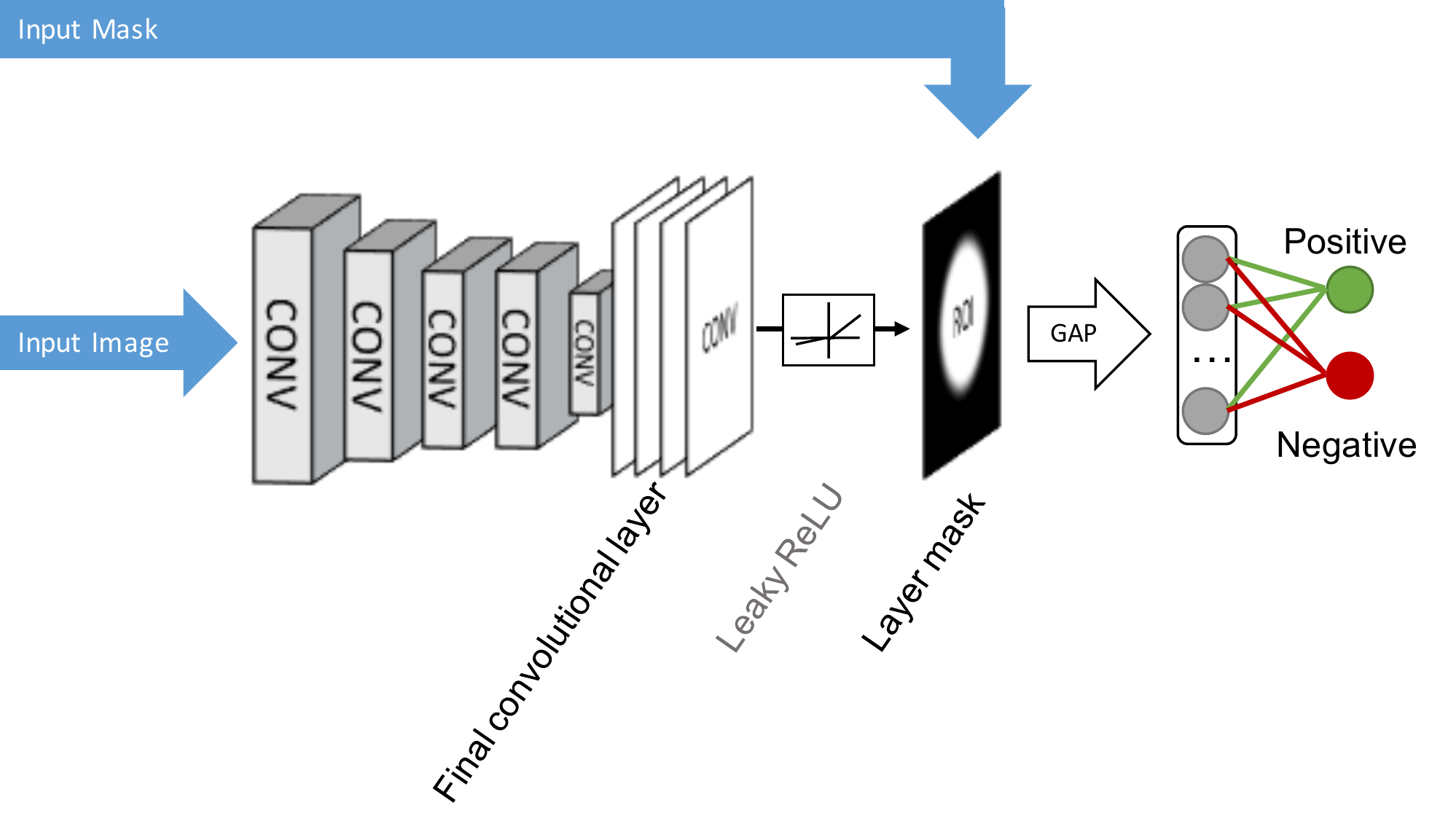}
    \caption{Localized learning by way of masking after the final convolutional layer. Each activation map in this layer, after activation through a leaky ReLU, is multiplied element-wise with a mask of the same size. This focuses gradient descent on particular parts of the image.}
    \label{fig:synapselearning}
\end{figure}

To resolve this problem, we introduce \textit{localized learning} to the GoogLeNet+GAP model~\cite{Zhou2016}. The na\"ive approach would be to obstruct parts of the image before input into the CNN. However, the shape of these obstructions would then become visible features which the CNN would recognize---an undesirable side affect. In the Global Average Pooling models, the \textit{final} convolutional layer contains locality information, which is then fed into a global average pooling (GAP) layer and a fully connected layer for classification purposes. By obstructing the \textit{activations of the final convolutional layer} (Figure~\ref{fig:synapselearning}), before the GAP layer, we force the fully-connected layer to only take into account activations found in the region-of-interest---this is essentially the inverse of class activation mapping. This masking \textit{guides the gradient descent as well}, forcing gradients outside of the region-of-interest towards zero. The final ReLU, Rectified Linear Unit,  layer is leaky with $\alpha = 0.1$ to mitigate the effect of dead ReLUs from overzealous masking.
 
Each training image is now paired with two training masks, a positive mask denoting its region of interest, and a negative mask denoting the background. For the purposes of online learning, the mask is created based on user input. Assuming the user clicks on point $x_c, y_c$ on a 14x14 grid, the positive layer mask Z is described by:

\begin{equation}
Z_{x,y} = \left\{
        \begin{array}{ll}
            G(x,y) & \quad  (x-x_c)^2 + (y-y_c)^2 > r^2 \\
            1 & \quad (x-x_c)^2 + (y-y_c)^2 \leq r^2
        \end{array}
    \right.
\end{equation}
where
\begin{equation}
G(x,y) = \exp{\left(-4\ln(2)\frac{\sqrt{(x-x_c)^2 + (y-y_c)^2}}{(1.5r)^2}\right)}
\end{equation}
and $r$ is selected such that a circle of radius $r$ on the mask layer will cover the target object. In our case, $r=4$ is selected conservatively. The softened circle further lessens the chance of dead ReLUs within the network. A negative mask is simply then $1-Z$. The entire mask, positive or negative, is then scaled such that the mean across it is 1. 

Each user click, then, defines not one training example, but \textit{two}: one positive example with the positive mask, and one negative example with the negative mask, with the same input image. This greatly reduces the need to seek out negative examples to reduce false positives. 

In contrast to object detection approaches, such as R-CNN~\cite{renNIPS15fasterrcnn} and YOLO~\cite{Redmon2015}, the masking approach presented here is still a classifier at its core, and can be easily extended to non-detection applications presented by~\cite{Zhou2016}, such as room classification or activity recognition. In our application, we forsee UAS being asked to detect, for instance, a hazardous situation or damaged area during a disaster response scenario. 

\subsection{Optical Flow for Training Assistance}

\begin{figure} \centering
    \includegraphics[width=\linewidth]{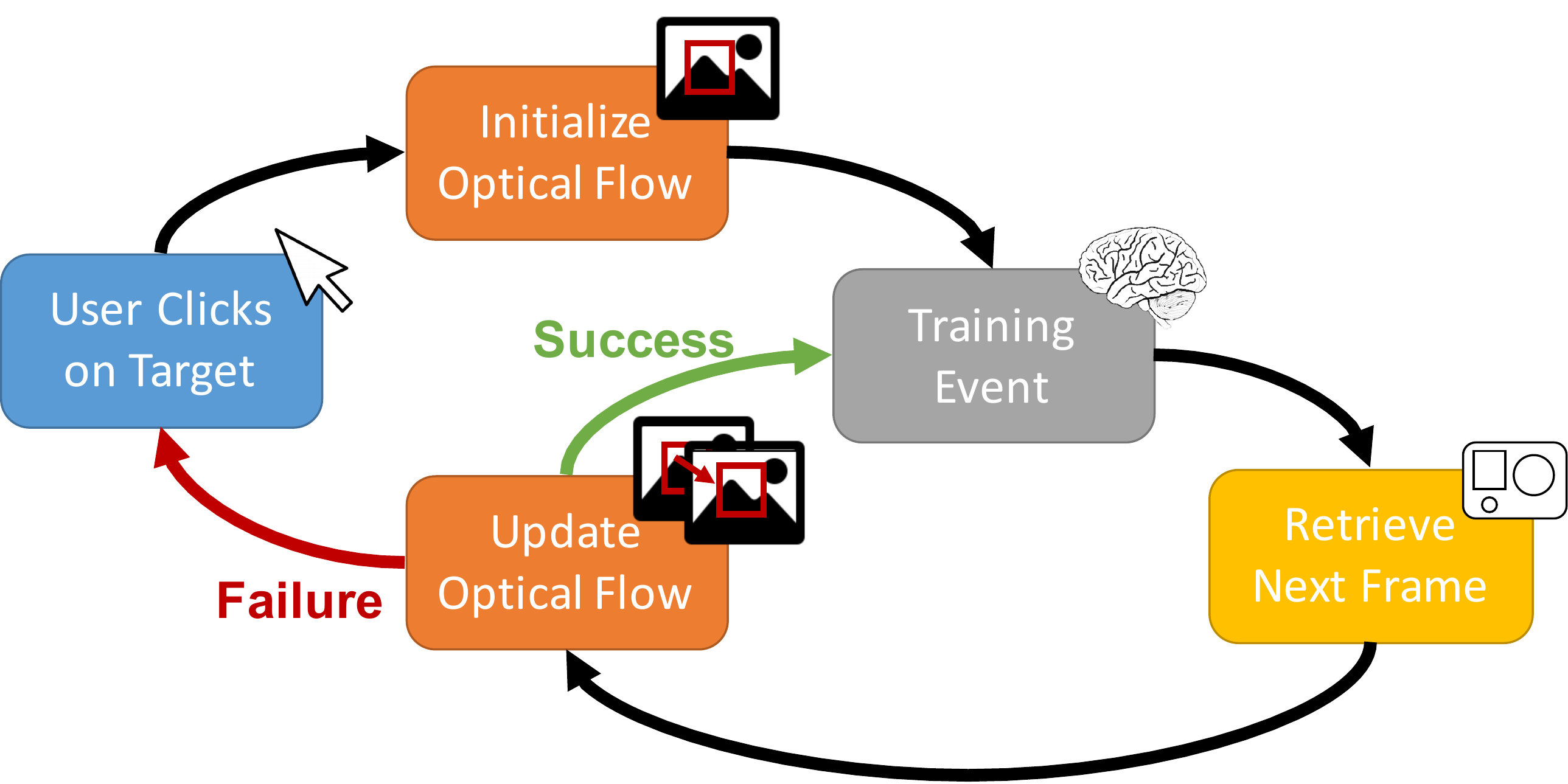}
    \caption{State diagram of optical flow-informed localized learning. Optical flow is initiated by a user interaction denoting the location of the target, and continues generating new training events as long as the tracking algorithm is tracking the target.}
    \label{fig:opticalflow}
\end{figure}

To further reduce required user interactions, we exploit a property particular to ToOT sequences: the target object is unlikely to have moved far between frames. We can thus assist the user with object tracking through optical flow, as long as the target item is in the scene. This tracked information, when coupled with foreground-background localized learning, can generate training events beyond what was produced by the user. Figure~\ref{fig:opticalflow} shows the method by which optical flow informs localized learning. When the user clicks on the target in the frame, an optical flow tracker is initialized on the target with a bounding box centered on the click. This initiates a training event with both the image and its background (masked). After training, a subsequent frame is retrieved from the camera, and the optical flow is updated using this frame. If the tracking update succeeds, another training event is initiated with the new location, and the cycle continues. If it fails, the user is asked to click on the target again.

In our experiments, we use the Median Flow~\cite{kalal2010forward}, because of its robust tracking failure detection. Because the consequences of a false positive tracking event is an improperly trained model and a noisy training set, we prefer an algorithm that plays on the safe side---however, any optical flow algorithm could be substituted here. 

\section{Experimental Setup}
\label{sec:experiments}
\subsection{Standardized User Behavior}

In field use, the effectiveness of our system is dependent on human input, \ie{} how accurately and often the operator conducts their tagging of the video frames, and how well the pilot tracks the target. In order to validate our approach and compare online training methodologies systematically, we construct two scenarios based on realistic use cases, each with \emph{train} and \emph{test} phases. Each phase consists of a time-ordered series of human-labeled images taken from the UAS in flight. In this way, we can replay training and test missions consistently and automatically. 

Each frame is tagged with two pieces of information: whether or not the target is in the frame and the coordinates of the center of the target if found. Using these pieces of information, we can simulate, in a repeatable fashion, a user supervising the learning process.


\subsection{Benchmark Datasets for Time-Ordered Online Training}

Existing datasets used in image classification and localization, such as the general-purpose ImageNet~\cite{Deng09imagenet} and the detection/segmentation Caltech-101~\cite{FeiFei2004} provide a rich corpus of tagged images of disparate scenes, but do not adequately represent  a single training session in the field, where each image in the set directly follows the last. The KITTI suite~\cite{Geiger2012CVPR} provides annotated video sequences, but they do not track any one particular training target, instead annotating general groups of items (\eg{} cars). In order to test our training system in the environment for which it was intended, \ie{} online training on a contiguous series of images captured during a UAS flight, we set up two scenarios with two different environments. 

We will refer to our scenarios as \emph{CarChaser} and \emph{PersonFinder}, where the target in question is a car or a person, respectively. Table~\ref{tab:datasets} describes the datasets in more detail. An emphasis was placed on having both positive and negative examples from the same scene. The test scenarios are intentionally less controlled, and re-enact a sequence where a UAS is attempting to find a hidden or moving target. For classification accuracy evaluation, the test sequence was pared down so that the number of positive and negative examples were equal. 

\begin{table}
\begin{center}
  \begin{tabular}{ | c | c | c | c | c |}
    \hline
     & \multicolumn{2}{c|}{Training Set} & \multicolumn{2}{c|}{Test Set} \\ \hline
    \textbf{Scenario} & \# Pos.  & $\sum_{x=1}^{f} u_x $ & \# Pos. & \# Neg. \\ \hline
    CarChaser & 397 & 323 & 129 & 129 \\ \hline
    PersonFinder & 188 & 182 & 85 & 85 \\ \hline
  \end{tabular}
   \caption{Number of images in training and test set for each scenario for PersonFinder and CarChaser. Test set is pared down to have equal positive and negative examples, so that classification accuracy can be used as a metric.}
   \label{tab:datasets}
  \end{center}
\end{table}

\subsubsection{CarChaser}

\begin{figure}[h]
  \includegraphics[width=\linewidth]{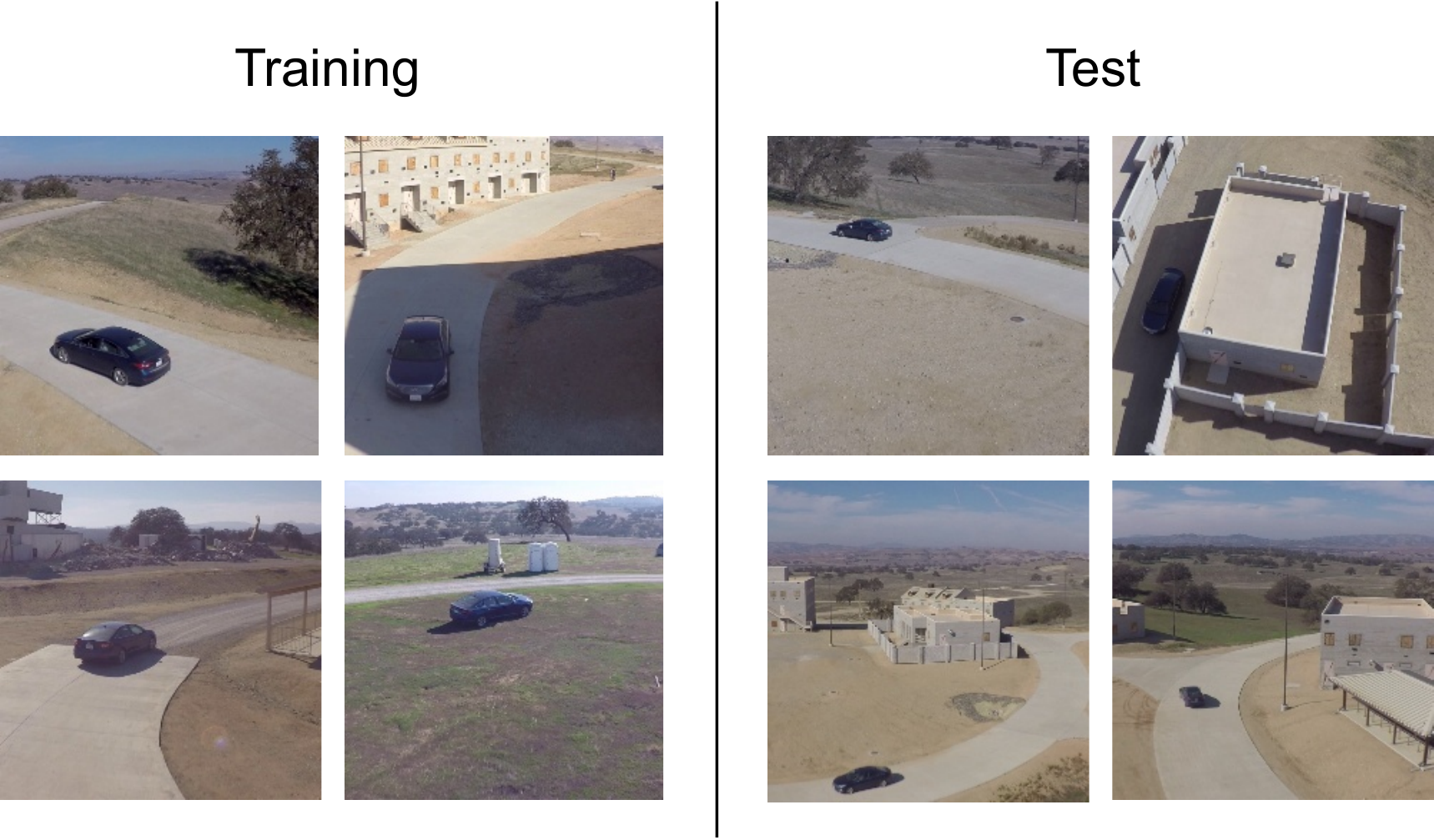}
    \caption{Sampled positive examples from the CarChaser training and test datasets.}
    \label{fig:carchaserset}
\end{figure}

In this scenario, the target is a blue sedan. The training portion of the scenario contains two ToOT sequences, in which the car moves around a circular road, stopping at certain points, giving the camera view of different backgrounds. Each scene also includes shots of the same background without the vehicle in view to use as negative examples. In the single test sequence, the car actively tries to evade the drone while the pilot attempts to catch up. Figure~\ref{fig:carchaserset} shows sample scenes from the training and test set. 

\subsubsection{PersonFinder}

\begin{figure}[h]
  \includegraphics[width=\linewidth]{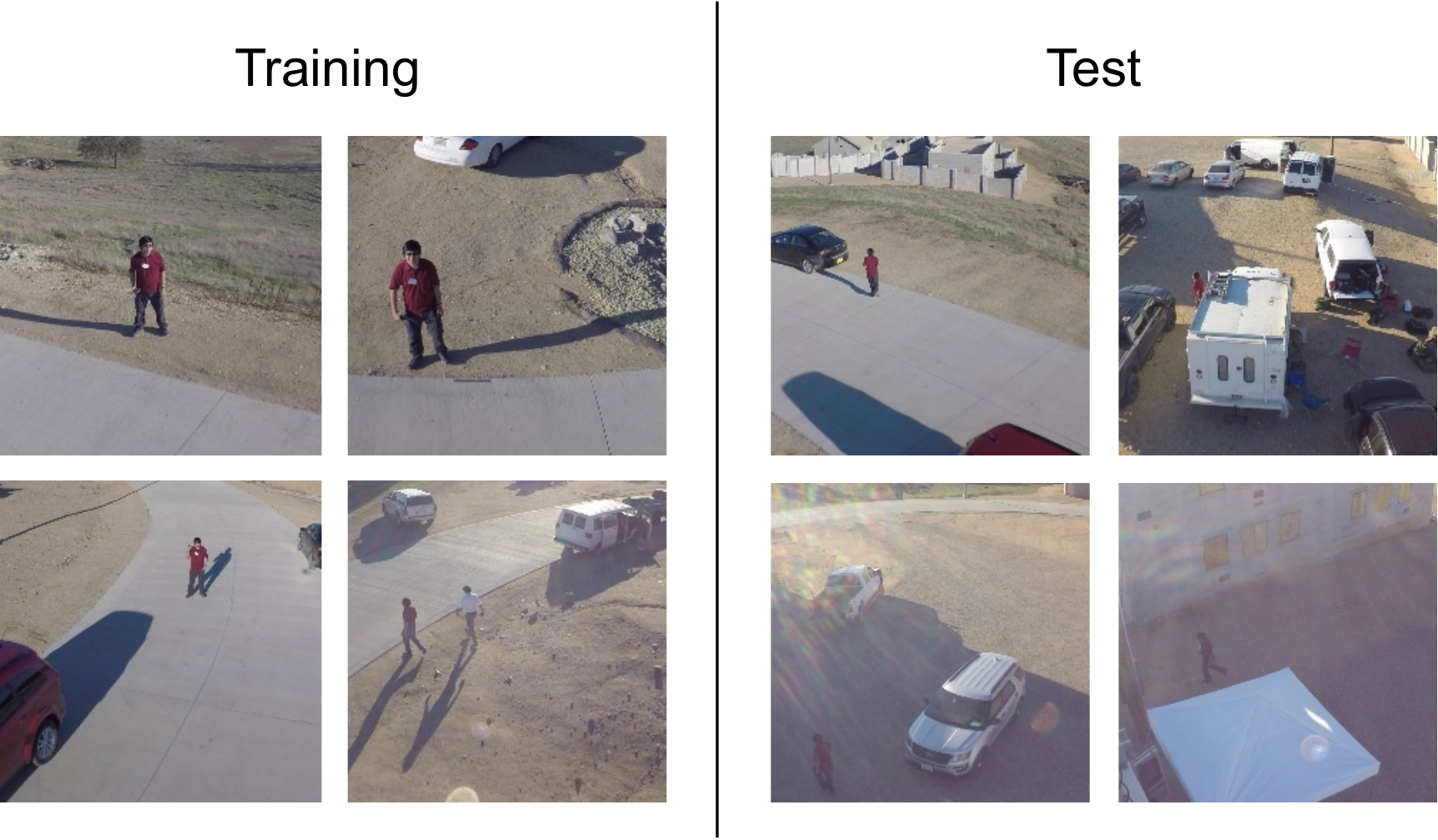}
    \caption{Sampled positive examples from the PersonFinder training and test datasets. Note the differences in size of the target relative to the frame and the lighting conditions, making this a much more difficult scenario than CarChaser.}
    \label{fig:personfinderdataset}
\end{figure}

In this scenario, our target is a particular person wearing a distinct red shirt (Figure~\ref{fig:personfinderdataset}). During the single training sequence, the person moves in and out of the frame against a simple background, and in the presence of other objects (vehicle, person wearing a white shirt). As with CarChaser, each background is featured with and without the target to facilitate training on negative examples. The training sequence consists of the person evading the UAS amongst various parked vehicles. 

\subsection{Training Strategies}

We evaluated the following training strategies on our two ToOT scenarios. 

\subsubsection{Offline (mini-batch) training.} All images and tags are known a-priori and trained in mini-batches of $b=8$ for $n$ rounds, where $n$ is the length of vector $V$. This is not a ToOT training strategy, and is done for reference purposes.

\subsubsection{Semi-Online training.} In this strategy, $u_i = 1, where i 
={1,...,n}$. In other words, the user tags every available image either positive or negative, and it is processed as described in Section~\ref{sec:semionline}.

\subsubsection{Localized learning.} We now apply localized learning (Section~\ref{sec:localizedlearning}). $u_i = 1$ only if $v_i$ contains the object of interest, and 0 otherwise. For each $u_i$ two copies of the image, one with a positive mask, and one with a negative mask, are added to the batch. This step is a prerequisite to using optical flow/object tracking, since object tracking only applies to positively tagged frames. 

\subsubsection{Optical flow-enhanced localized learning.} In this strategy, we assist our standardized user using optical flow (OF), as described in Section~\ref{sec:localizedlearning}. We make the following assumptions about how the user uses the OF-assisted system:
\begin{itemize}
\item When no OF is initialized, and the target is in the scene, the user clicks on the target and initializes OF. 
\item When OF updates are successful, the user allows the system to perform training without interruption.
\item When OF tracking loses the target, the user again clicks on the target in the next frame, re-initializing the OF tracking. 
\end{itemize}

This means that on many occasions, $v_i$ is added to the mini-batch \textit{even if the corresponding} $u_i = 0$. This produces additional training benefit without additional user input. 

\subsection{Experiments}

In our experiments, we want to 1. show that semi-online learning is comparable in performance to offline mini-batch learning, and 2. quantify the benefit of using optical-flow assisted training for time-ordered online training of an image classifier. To do so, we will look at several different metrics. All tests use as a base the GoogLeNet-GAP model, pre-trained on ImageNet as provided by~\cite{Zhou2016}. Adadelta~\cite{Zeiler2012} is used as the gradient descent algorithm. To mitigate the stochasticity of the gradient descent algorithm, all metrics were computed from a mean of ten runs. 

\subsubsection{Maximum Accuracy}
\label{sec:maxacc_exp}

In real-world usage, we are less concerned with finishing the entire training sequence as we are with achieving the best possible accuracy, $A_{max}$ during training, where 

\begin{equation}
A_{max} = \max_{x=i}^{n} A_i
\end{equation}

In other words, $A_{max}$ is the highest achievable point when all frames to $n$ are used. We began by taking $A_{max}$ of offline (traditional mini-batch learning) with $b=8$, and used this as a baseline for evaluating the other measurements. We then compared this $A_{max}$ to semi-online training to determine that it was not a detriment to the performance of the resulting model.


\subsubsection{CTB and ITB}

In Section~\ref{sec:background}, we defined $CTB$ and $ITB$ as metrics particular to ToOT that are a measure of the \textit{effectiveness} of each user interaction. In particular, $\overline{ITB}_{(i,j)}$ can be used to compare training strategies, given all achieve a fixed final accuracy $A_f$. We assigned $A_f$ to be the minimum $A_{max}$ of the various training strategies. Runs were terminated after $A_f$ was reached. 

We plotted $CTB_{u_i}$ and $ITB_{u_i}$ over $v_i$, as well as computed $\overline{ITB}_{(1,f)}$, where $f$ is the step when $A_f$ is reached for that particular scenario.

\section{Experimental Results}
\label{sec:results}
\subsection{Maximum Accuracy}

We compare the achieved $A_{max}$, after all training images have been used, between the following training strategies:
\begin{itemize}
\item offline (mini-batch) learning, $b=8$
\item online learning, $b=8$
\item localized learning with $b=2$ 
\item localized learning with $b=8$
\item OF-assisted localized learning with $b=2$
\item OF-assisted localized learning with $b=8$
\end{itemize}

\begin{figure} \centering
    \includegraphics[width=\linewidth]{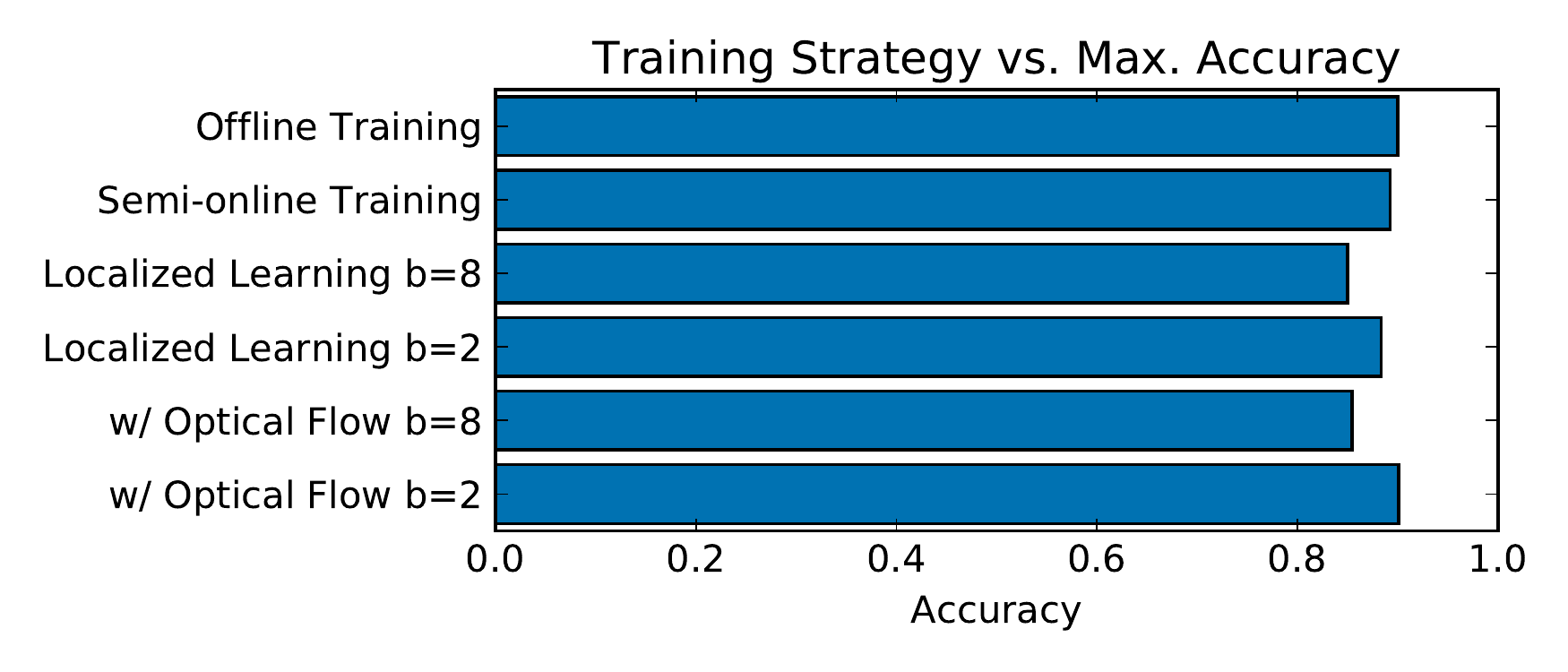}
    \caption{Comparison of $A_{max}$ between training strategies for CarChaser scenario. Offline and semi-online training produce very similar results, with the other methods also being comparable. Smaller batch sizes for localized learning with and without optical flow produce slightly higher accuracies.}
    \label{fig:carchaser_acc}
\end{figure}

\begin{figure} \centering
    \includegraphics[width=\linewidth]{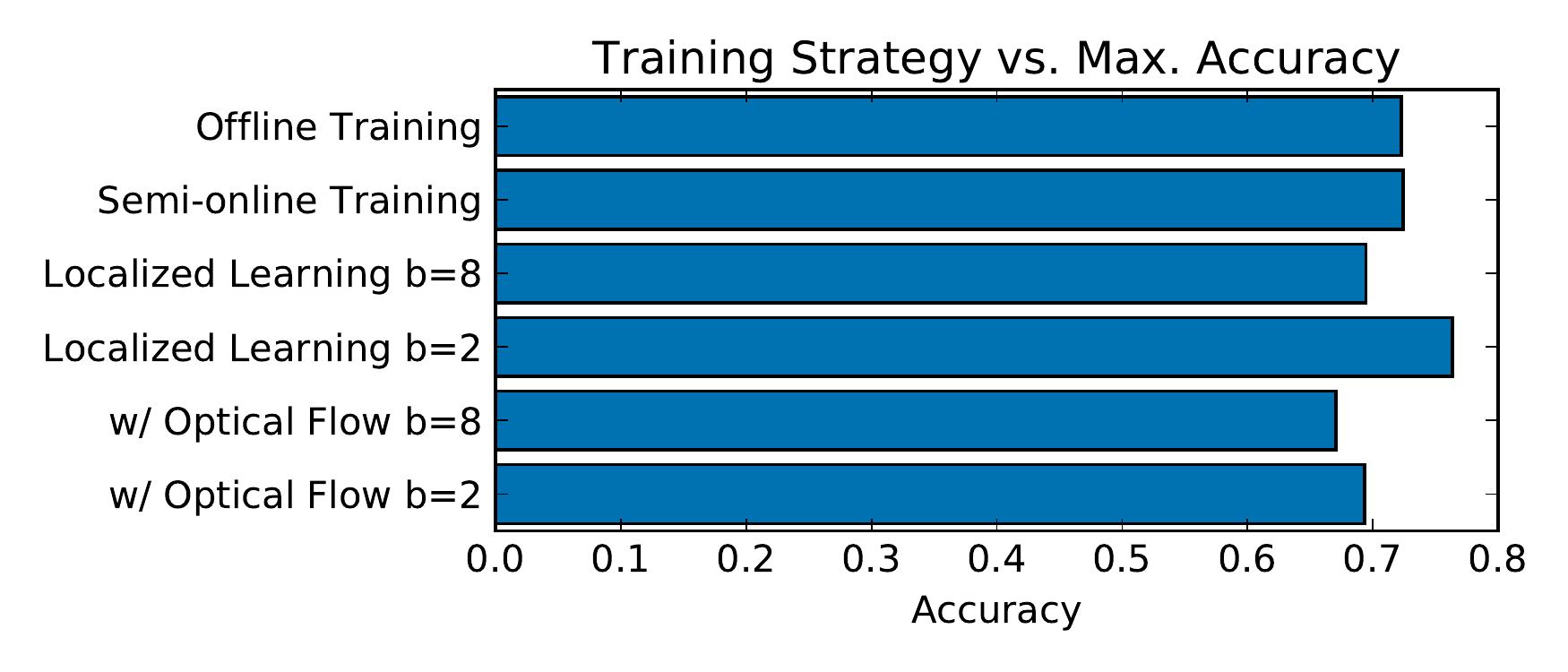}
    \caption{Comparison of $A_{max}$ between training strategies for PersonFinder scenario. Offline and semi-online training produce very similar results, with the other methods also being comparable. Note that this scenario's test set is significantly more difficult (much smaller target) and smaller than the CarChaser scenario, and the $A_{max}$ is smaller overall.}
    \label{fig:personfinder_acc}
\end{figure}

Figure~\ref{fig:carchaser_acc} shows this comparison for the CarChaser scenario, and Figure~\ref{fig:personfinder_acc} for the PersonFinder scenario. We see that online learning performs comparably to offline learning, showing that the incremental introduction of new data and tags into the system is not detrimental to the overall performance of the model, given a fixed dataset. 

Localized learning and OF-assisted localized learning do not always achieve the same accuracy as the other two methods---for CarChaser, the accuracy is slightly higher, and for PersonFinder, the accuracy is slightly lower. This is logical, since both these methods use only the positively tagged subset of the available imagery---they do not have, strictly speaking, the same training set. Despite this, we see reasonable performance without having any negative examples.

\subsection{Efficacy of User Interactions}

\subsubsection{CarChaser}

\begin{figure*}
\centering
\begin{minipage}[t]{.49\textwidth}
    \includegraphics[width=\textwidth]{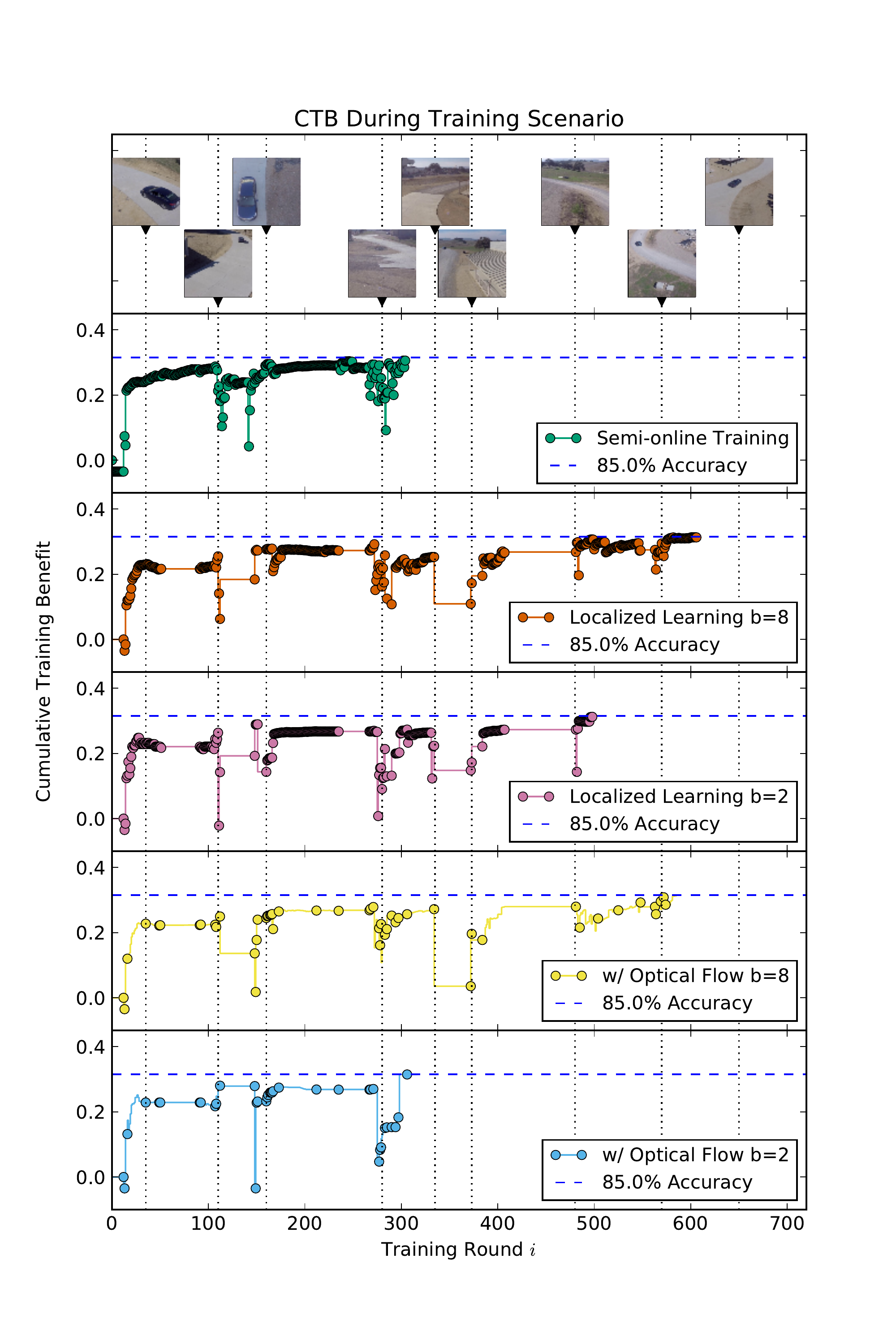}
	\caption{Plot of cumulative training benefit for CarChaser scenario. Dots on the traces represent user interactions. Horizontal blue line represents $A_f$. We note that the number of video frames (\ie{} time) required for the semi-online training and optical flow with $b=2$ is comparable.}
    \label{fig:carchaser_ctb}
\end{minipage}\hfill
\begin{minipage}[t]{.49\textwidth}
    \includegraphics[width=\textwidth]{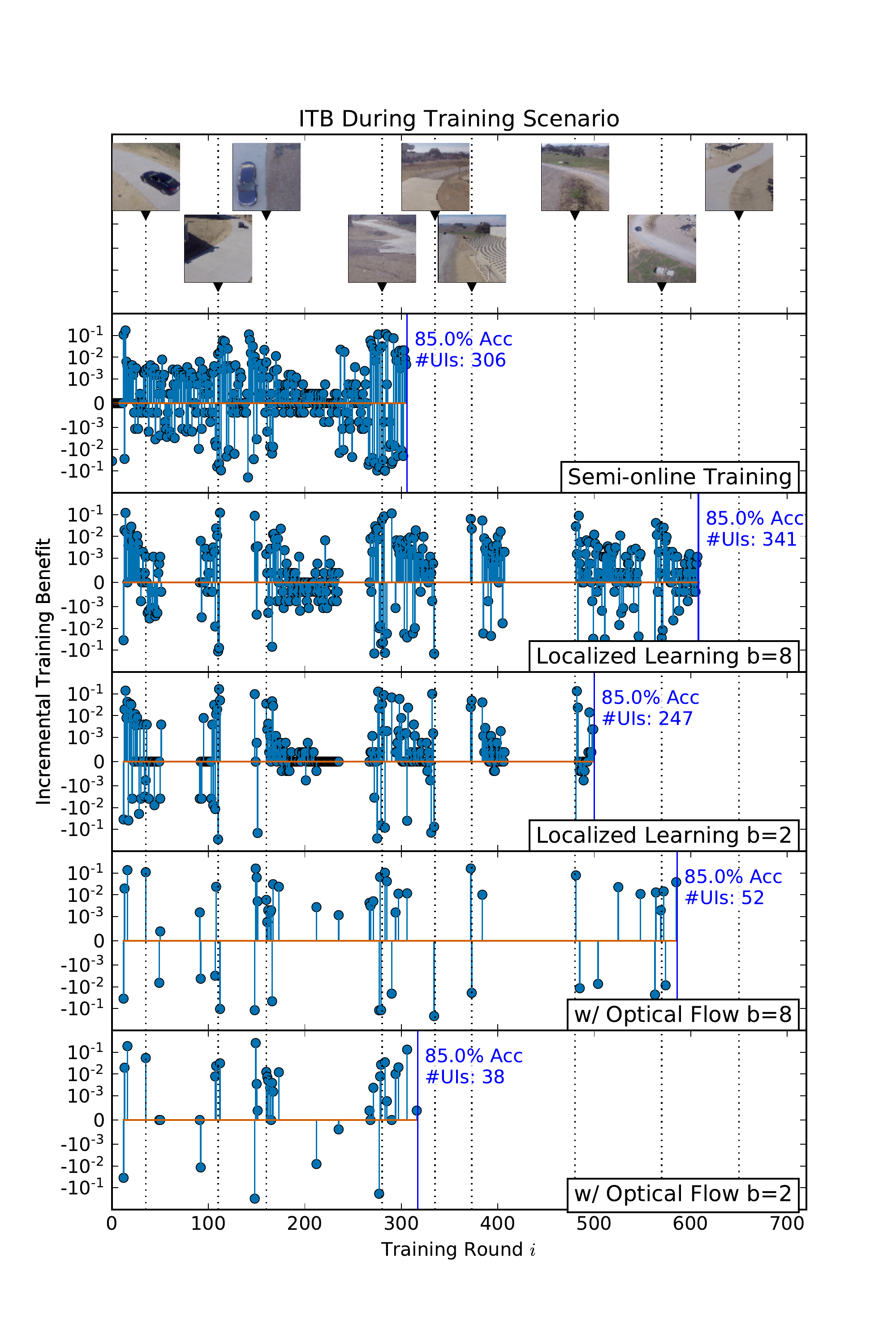}
    \caption{Plot of incremental training benefit for CarChaser scenario. Each point represents a user interaction. Vertical blue line represents point where accuracy reaches $A_f$. We see that the optical flow-enabled methods require far less user interactions before $A_f$ is met, although they take roughly the same length of video.}
    \label{fig:carchaser_itb}
\end{minipage}
\end{figure*}

For CarChaser, the final accuracy $A_f =  0.85$. In Figure~\ref{fig:carchaser_ctb}, we plot the CTB of all the aformentioned training strategies up to $A_f$. The top plot shows representative images in the training scenario at that period of time. Each trace follows the accuracy at each training round $i$, regardless of if a user interaction happens at that step ($u_i = 1$). Each dot on the trace represents the rounds in which a user interaction did happen. 

Flat regions in the localized learning and OF-enabled localized learning plots represent periods of no action, \ie{}  places where the target is not in the scene and the frames are not being trained on. Note that the lines on the ``with OF'' plots closely follow those of the localized learning plots---but the number of dots (user interactions) is much less dense. This means that, as expected, optical flow enables the generation of user interactions that are \textit{not} triggered by the user but still effective in training the model.

We note several other interesting trends in these plots. We also notice troughs in $PI/UI$ that do not match the general falloff trend; these correspond to parts of the training sequence where the training actually \textit{worsens} the accuracy. The two largest troughs correspond to parts in the training sequence where the target car is small relative to the frame. This suggests that certain properties of the image may be predictors of their effectiveness in training the model---we leave this question for future study. 

Figure~\ref{fig:carchaser_itb} plots the incremental training benefit (ITB) for each user interaction in Figure~\ref{fig:carchaser_ctb}. As expected, the OF-enabled methods produce far less dense plots as they have far fewer user interactions (Table~\ref{tab:uis}).

We note that while some of the user interactions produce a very large or very small (negative) $ITB$, the vast majority produce very small changes. To emphasize the smaller values, the y-axis is plotted on a logarithmic scale. Semi-online training shows this trend the strongest, with many, many low-impact user interactions triggering low-impact training rounds, each contributing to the final accuracy. With the optical flow-enabled methods, these low-impact training rounds can be triggered several at a time.

\begin{figure} \centering
    \includegraphics[width=\linewidth]{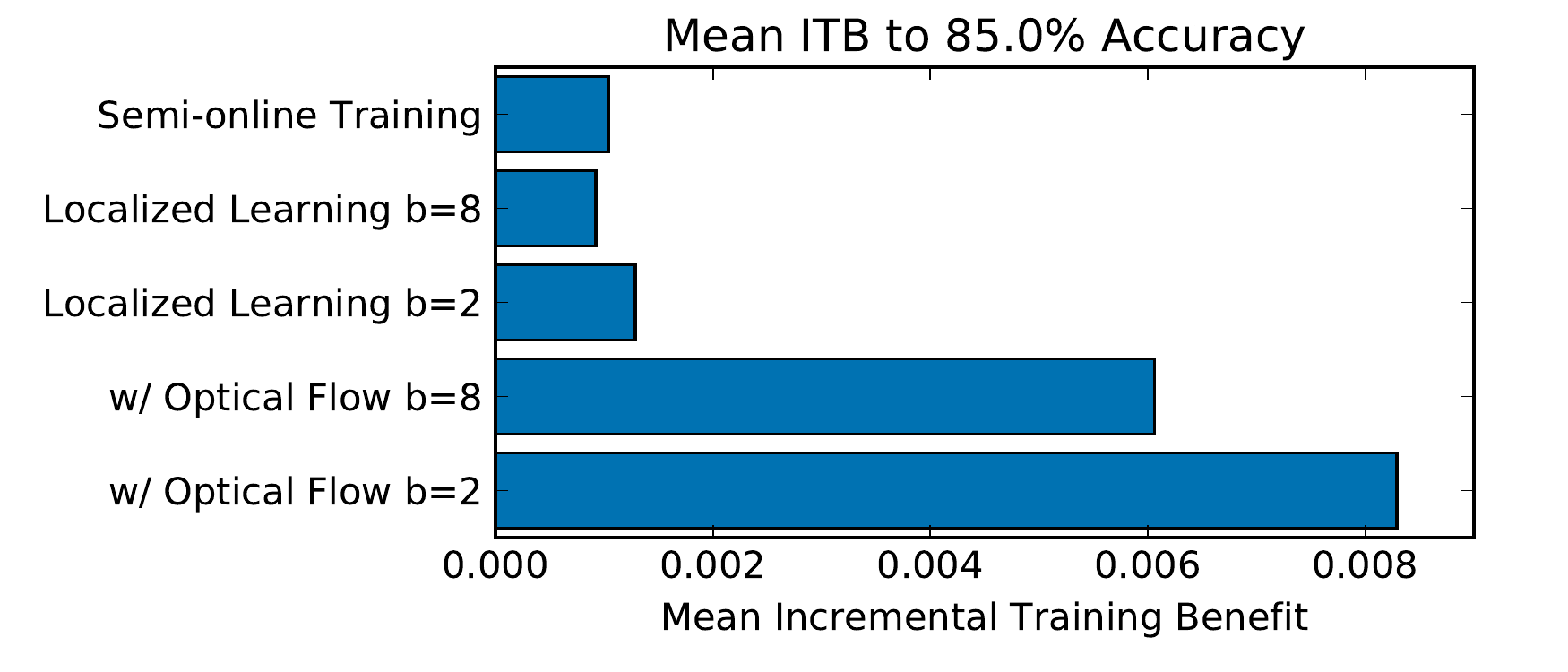}
    \caption{Graph of mean incremental training benefit for CarChaser scenario, up to $A_f = 0.85$. Both optical flow methods produce higher mean \textit{ITB} than without, with a batch size of 2 producing the highest.}
    \label{fig:carchaser_meanitb}
\end{figure}

The mean $ITB$, $\overline{ITB}_{(1,f)}$,for each training technique is shown in Figure~\ref{fig:carchaser_meanitb}---the average of every point in Figure~\ref{fig:carchaser_itb}.

\begin{figure*}
\centering
\begin{minipage}[t]{.49\textwidth}
    \includegraphics[width=\textwidth]{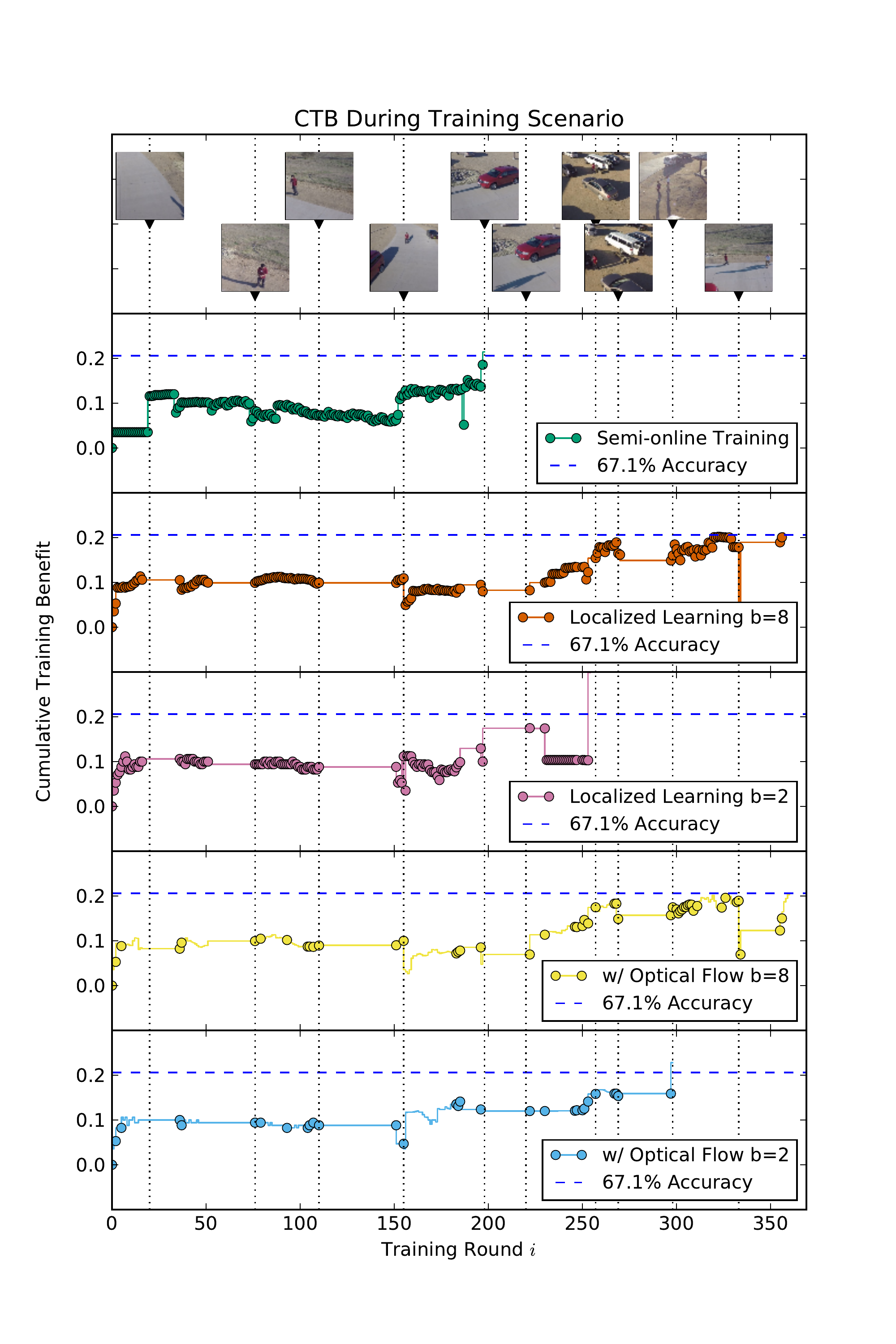}
	\caption{Plot of cumulative training benefit for PersonFinder scenario. Dots on the traces represent user interactions. Horizontal blue line represents $A_f$. We note that the number of video frames (\ie{} time) required for the semi-online training and optical flow with $b=2$ is comparable.}
    \label{fig:personfinder_ctb}
\end{minipage}\hfill
\begin{minipage}[t]{.49\textwidth}
    \includegraphics[width=\textwidth]{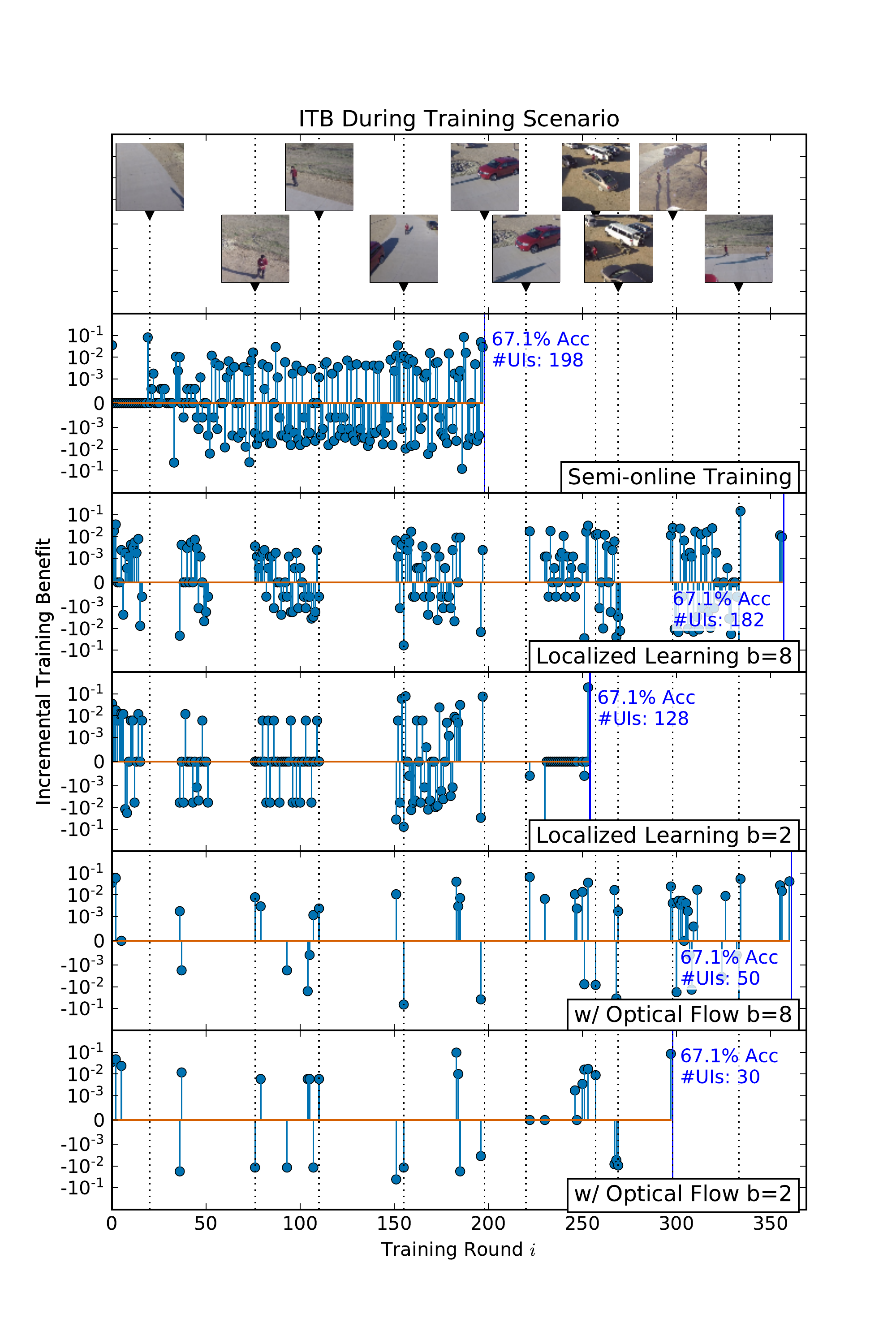}
    \caption{Plot of incremental training benefit for PersonFinder scenario. Each point represents a user interaction. Vertical blue line represents point where accuracy reaches $A_f$. We see that the optical flow-enabled methods require far less user interactions before $A_f$ is met, although they take roughly the same length of video.}
    \label{fig:personfinder_itb}
\end{minipage}
\end{figure*}

We see that the optical-flow enabled strategy enables user interactions that are more than 6 times more effective (\ie{} produce on average 6 times more $ITB$ per interaction) than semi-online training when $b=8$, and 8 times when $b=2$. 

\subsubsection{PersonFinder}

We repeated the same experiments on the PersonFinder dataset. For PersonFinder, which had a more difficult test scenario and a smaller target, $A_f = 0.671$.

In Figure~\ref{fig:personfinder_ctb}, we see the same effect of localized learning and localized learning with optical flow, vs. semi-online learning. All plots show an interesting trend, however, that the $CTB_i$ did not climb significantly until around $i=150$. 

We see the same reflected in the $ITB_{u_i}$ plot, with the user interactions that happened earlier having a smaller $ITB_{u_i}$ than the ones that happen later. This implies that those early training examples were less ``useful'' than those that happen later, suggesting yet again that certain images---in this case images with very sparse backgrounds---are less valuable than others. 

\begin{figure}
\centering
    \includegraphics[width=\linewidth]{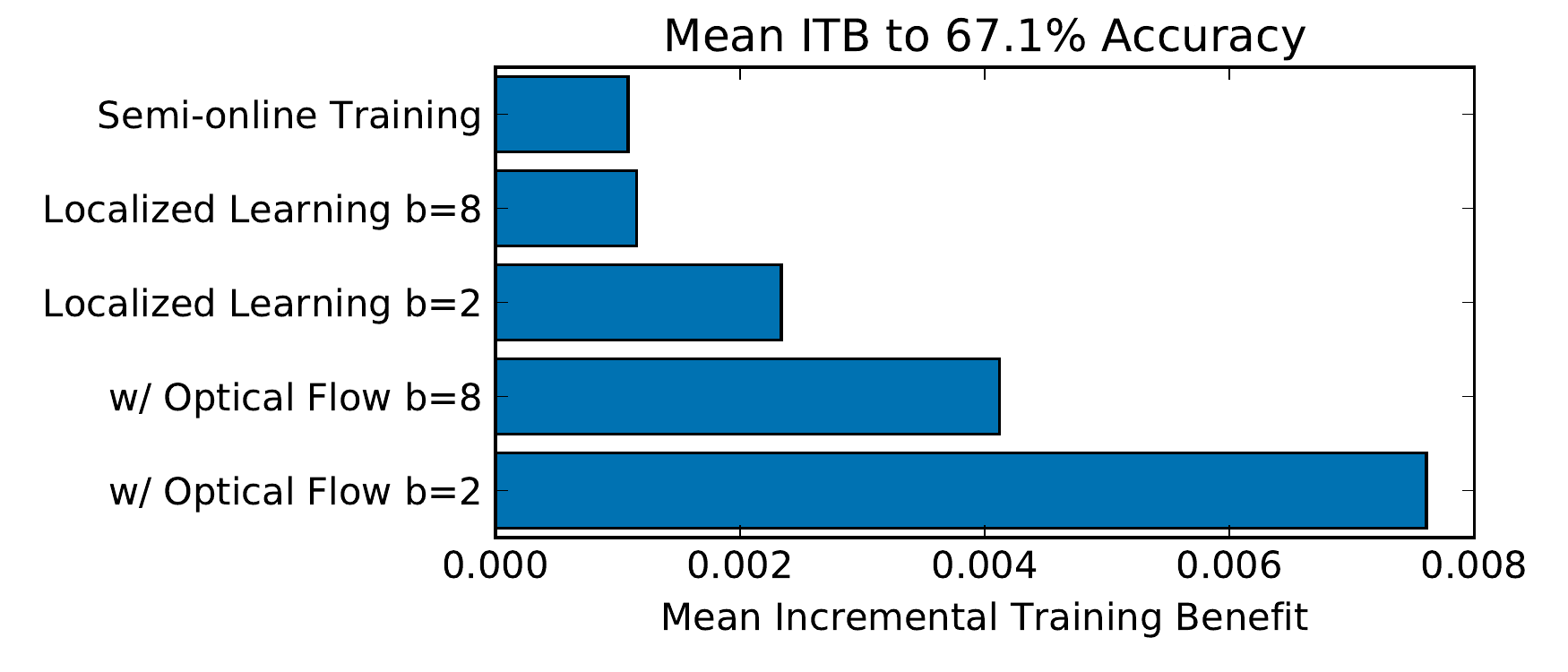}
    \caption{Graph of mean incremental training benefit for PersonFinder scenario, up to $A_f = 0.671$. Both optical flow methods produce higher mean \textit{ITB} than without, with a batch size of 2 producing the highest.}
    \label{fig:personfinder_meanitb}
\end{figure}

Figure~\ref{fig:personfinder_meanitb} again plots $\overline{ITB}_{(1,f)}$ for the various training strategies. From the perspective of reducing user interactions, localized learning with optical flow again dramatically increased $\overline{ITB}_{(1,f)}$, 5 times more than semi-online training when $b=8$, and 8 times when $b=2$. Table~\ref{tab:uis} shows both the total user interactions required to reach $A_f$, as well as the mean $ITB_{u_i}$ value for these interactions. 

We notice another interesting trend from this data---that not only does adding optical flow increase $\overline{ITB}_{(1,f)}$, so does \textit{reducing the batch size}. While a small batch size could cause instability---especially when all examples within the batch are of the same label, as would be common with video sequences---having both the positively masked portion of the image and the negatively matched portion of the image in the same batch mitigates this effect. This is in accordance with the findings of~\cite{Wilson2003}, adapted to the ToOT case.

\begin{table}[H]
\begin{center}
  \begin{tabular}{ | c | c | c | c | c |}
    \hline
     & \multicolumn{2}{c|}{CarChaser} & \multicolumn{2}{c|}{PersonFinder} \\ \hline
    \textbf{Strategy} & $\sum_{i=1}^{f} u_i $& $\overline{ITB}_{(1,f)}$ & $\sum_{i=1}^{f} u_i $& $\overline{ITB}_{(1,f)}$ \\ \hline
    Semi-online & 306 & $1.04 * 10^{-3}$ & 198 &$ 1.08 * 10^{-3}$  \\ \hline
    Local. $b=8$ & 341 & $9.24* 10^{-4}$ & 182 & $1.15 * 10^{-3}$ \\ \hline
    Local. $b=2$ & 247 &  $1.28*10^{-3}$ & 128 &  $2.33 * 10^{-3}$   \\ \hline
    w/ OF $b=8$ & 51 & $6.05*10^{-3}$ & 50 & $4.12 * 10^{-3}$   \\ \hline
    w/ OF $b=2$ & 37 & $8.29*10^{-3}$ & 30 & $7.61 * 10^{-3} $  \\ \hline
  \end{tabular}
   \caption{Number of user interactions and mean $ITB$ for the different training strategies. We see that the combined approach of optical flow and localized learning is 5-8 times more effective than semi-online.}
   \label{tab:uis}
  \end{center}
\end{table}

\section{Conclusion}
\label{sec:conclusion}


While recent work has exemplified the power of CNN-based algorithms for flexible, robust image classification and recent hardware advances have made them feasible to deploy in embedded systems, these algorithms typically require offline pre-training before deployment. In this work, we recognize that training CNN classifiers and object detectors in real-time via human input encompasses a unique class of problem, which we refer to as Time-ordered Online Training. In such problems, we are not only concerned with the accuracy of the model produced, but also the human effort, \ie{} the number of user interactions, required to train the model. We define \textit{training benefit} as a metric for measuring the impact of user interactions, and use the \textit{mean incremental training benefit} ($\overline{ITB}_{(1,f)}$), given a fixed final accuracy $A_f$, to compare various training strategies in the effectiveness of each user interaction. We show that by exploiting the time-ordered nature of our video stream through localized learning combined with optical flow object tracking, we can multiply mean incremental training benefit by about 8 times over the one-frame-one-tag approach. 

We believe this work sets the stage for further investigation into real-time training on ToOT sequences. Optical flow tracking is just one example of how the properties of time-ordered video streams can be used to reduce user effort. From Figures~\ref{fig:carchaser_itb} and \ref{fig:personfinder_itb}, we can observe that while optical flow effectively reduces the number of user interactions required to reach $A_f$, it does not eliminate low or even negative $ITB_{u_i}$ interactions. Further work in information \textit{extraction} from each frame as well as in frame \textit{selection} could mitigate this problem. The addition of multiple assistive functions will also necessitate a model where multiple inputs, of which the human user is one, are used to train the CNN collaboratively. 

\section{Acknowledgements}

We would like to thank Cef Ramirez, whose work on the Telemetry DB was an enabler for this work, and also for appearing in the training sequences. We would also like to thank the Naval Postgraduate School for their continued support of the JIFX event, without which these studies would not be possible. 

\bibliography{Untitled.bib}

\begin{thebibliography}{10}

\bibitem{Guisti2016}
{A Machine Learning Approach to Visual Perception of Forest Trails for Mobile
  Robots}.
\newblock {\em Robotics and Automation Letters}, 1(2):661--667, 2016.

\bibitem{Abadi2015}
M.~Abadi et~al.
\newblock {TensorFlow: Large-Scale Machine Learning on Heterogeneous
  Distributed Systems}.
\newblock 2015.

\bibitem{Beauchemin:1995}
S.~S. Beauchemin and J.~L. Barron.
\newblock The computation of optical flow.
\newblock {\em ACM Comput. Surv.}, 27(3):433--466, Sept. 1995.

\bibitem{Bojarski2016}
M.~Bojarski, D.~{Del Testa}, D.~Dworakowski, B.~Firner, B.~Flepp, P.~Goyal,
  L.~D. Jackel, M.~Monfort, U.~Muller, J.~Zhang, X.~Zhang, J.~Zhao, and
  K.~Zieba.
\newblock {End to End Learning for Self-Driving Cars}.
\newblock {\em arXiv:1604}, pages 1--9, 2016.

\bibitem{bottou2016}
L.~{Bottou}, F.~E. {Curtis}, and J.~{Nocedal}.
\newblock {Optimization Methods for Large-Scale Machine Learning}.
\newblock {\em ArXiv e-prints}, June 2016.

\bibitem{Deng09imagenet}
J.~Deng, W.~Dong, R.~Socher, L.~jia Li, K.~Li, and L.~Fei-fei.
\newblock Imagenet: A large-scale hierarchical image database.
\newblock In {\em CVPR}, 2009.

\bibitem{FeiFei2004}
L.~Fei-Fei, R.~Fergus, and P.~Perona.
\newblock Learning generative visual models from few training examples: An
  incremental bayesian approach tested on 101 object categories.
\newblock In {\em 2004 Conference on Computer Vision and Pattern Recognition
  Workshop}, pages 178--178, June 2004.

\bibitem{Geiger2012CVPR}
A.~Geiger, P.~Lenz, and R.~Urtasun.
\newblock Are we ready for autonomous driving? the kitti vision benchmark
  suite.
\newblock In {\em Conference on Computer Vision and Pattern Recognition
  (CVPR)}, 2012.

\bibitem{Girshick2014}
R.~Girshick, J.~Donahue, T.~Darrell, and J.~Malik.
\newblock {Rich feature hierarchies for accurate object detection and semantic
  segmentation}.
\newblock {\em Proceedings of the IEEE Computer Society Conference on Computer
  Vision and Pattern Recognition}, pages 580--587, 2014.

\bibitem{Grauman2010}
K.~Grauman and B.~Leibe.
\newblock {\em {Visual Object Recognition}}, volume~5.
\newblock 2010.

\bibitem{Kading2016}
C.~K{\"{a}}ding, E.~Rodner, A.~Freytag, and J.~Denzler.
\newblock {Fine-tuning Deep Neural Networks in Continuous Learning Scenarios}.
\newblock In {\em ACCV 2016 Workshop on Interpretation and Visualization of
  Deep Neural Nets}, 2016.

\bibitem{kalal2010forward}
Z.~Kalal, K.~Mikolajczyk, and J.~Matas.
\newblock Forward-backward error: Automatic detection of tracking failures.
\newblock In {\em Pattern recognition (ICPR), 2010 20th international
  conference on}, pages 2756--2759. IEEE, 2010.

\bibitem{Kiswani2016}
A.~Kiswani, A.~Aides, and M.~Silberstein.
\newblock {Deep Learning in Aerial Systems Using Jetson}, 2016.

\bibitem{Krizhevsky2012}
A.~Krizhevsky, I.~Sutskever, and G.~E. Hinton.
\newblock {ImageNet Classification with Deep Convolutional Neural Networks}.
\newblock {\em Advances In Neural Information Processing Systems}, pages 1--9,
  2012.

\bibitem{LeCun2006}
Y.~LeCun, U.~Muller, J.~Ben, E.~Cosatto, and B.~Flepp.
\newblock {Off-road Obstacle Avoidance Through End-to-end Learning}.
\newblock {\em Advances in neural information processing systems}, 18:739,
  2006.

\bibitem{Porter2001}
S.~Porter, M.~Mirmehdi, and B.~Thomas.
\newblock {Detection and Classification of Shot Transitions}.
\newblock In {\em Procedings of the British Machine Vision Conference}, pages
  9.1--9.10, 2001.

\bibitem{Redmon2015}
J.~Redmon, S.~Divvala, R.~Girshick, and A.~Farhadi.
\newblock {You Only Look Once: Unified, Real-Time Object Detection}.
\newblock {\em eprint arXiv:1506.02640}, 2015.

\bibitem{renNIPS15fasterrcnn}
S.~Ren, K.~He, R.~Girshick, and J.~Sun.
\newblock Faster {R-CNN}: Towards real-time object detection with region
  proposal networks.
\newblock In {\em Advances in Neural Information Processing Systems ({NIPS})},
  2015.

\bibitem{Symington2010}
A.~Symington, S.~Waharte, S.~Julier, and N.~Trigoni.
\newblock {Probabilistic target detection by camera-equipped UAVs}.
\newblock {\em Proceedings - IEEE International Conference on Robotics and
  Automation}, 67:4076--4081, 2010.

\bibitem{Szegedy2014}
C.~Szegedy, W.~Liu, Y.~Jia, and P.~Sermanet.
\newblock {Going Deeper with Convolutions}.
\newblock {\em arXiv preprint arXiv: 1409.4842}, 2014.

\bibitem{Wilson2003}
D.~R. Wilson and T.~R. Martinez.
\newblock {The general inefficiency of batch training for gradient descent
  learning}.
\newblock {\em Neural Networks}, 16(10):1429--1451, 2003.

\bibitem{Yosinski2014}
J.~Yosinski, J.~Clune, Y.~Bengio, and H.~Lipson.
\newblock {How transferable are features in deep neural networks?}
\newblock {\em Advances in Neural Information Processing Systems 27
  (Proceedings of NIPS)}, 27:1--9, 2014.

\bibitem{Zeiler2012}
M.~D. Zeiler.
\newblock {ADADELTA: An Adaptive Learning Rate Method}.
\newblock {\em arXiv}, page~6, 2012.

\bibitem{Zhou2016}
B.~Zhou, A.~Khosla, A.~Lapedriza, A.~Oliva, A.~Torralba, A.~Iccv, and P.~Id.
\newblock {Learning Deep Feature for Discriminative Localization}.
\newblock {\em CVPR}, 2016.

\end{thebibliography}
\bibliographystyle{abbrv}
\end{document}